\documentclass{article}

\usepackage{arxiv}

\usepackage[utf8]{inputenc} 
\usepackage[T1]{fontenc}    
\usepackage[hidelinks]{hyperref}       
\usepackage{url}            
\usepackage{booktabs}       
\usepackage{amsfonts}       
\usepackage{nicefrac}       
\usepackage{microtype}      
\usepackage{lipsum}         
\usepackage{doi}

\usepackage[numbers]{natbib}
\usepackage{graphicx}
\usepackage{amsmath}
\usepackage{algorithm}
\usepackage{algorithmic}
\usepackage{here}
\usepackage{listings}
\usepackage{multirow}
\usepackage{booktabs}

\lstdefinestyle{mystyle}{
    basicstyle=\ttfamily\footnotesize,
    breakatwhitespace=false,         
    breaklines=true,                 
    captionpos=b,                    
    keepspaces=true,                 
    numbers=left,                    
    numbersep=1.5pt,                  
    showspaces=false,                
    showstringspaces=false,
    showtabs=false,                  
    tabsize=2
}

\title{Fully Data-driven but Interpretable Human Behavioural Modelling with Differentiable Discrete Choice Model}

\date{}

\author{\href{https://orcid.org/0000-0001-9247-4104}{\includegraphics[scale=0.06]{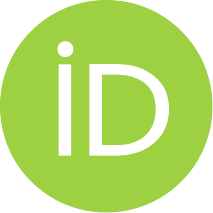}\hspace{1mm}Fumiyasu Makinoshima}\thanks{Corresponding author: Fumiyasu Makinoshima (f.makinoshima@fujitsu.com)} \\
	Fujitsu Limited\\
	Kawasaki\\
	Japan\\
	\texttt{f.makinoshima@fujitsu.com} \\
	\And
  Tatsuya Mitomi \\
	Fujitsu Limited\\
	Kawasaki\\
	Japan\\
	\texttt{tatsuya.mitomi@fujitsu.com} \\
  \And
  Fumiya Makihara \\
	Fujitsu Limited\\
	Kawasaki\\
	Japan\\
	\texttt{makihara.fumiya@fujitsu.com} \\
  \And
  Eigo Segawa \\
	Fujitsu Limited\\
	Kawasaki\\
	Japan\\
	\texttt{segawa.eigo@fujitsu.com} \\
}

\renewcommand{\shorttitle}{\textit{arXiv} Template}

\hypersetup{
pdftitle={Fully Data-driven but Interpretable Human Behavioural Modelling with Differentiable Discrete Choice Model},
pdfsubject={cs.MA, cs.LG},
pdfauthor={F. Makinoshima et al.},
}

\begin{document}
\maketitle

\begin{abstract}
  Discrete choice models are essential for modelling various decision-making processes in human behaviour.
  However, the specification of these models has depended heavily on domain knowledge from experts, and the fully automated but interpretable modelling of complex human behaviours has been a long-standing challenge.
  In this paper, we introduce the differentiable discrete choice model (Diff-DCM), a fully data-driven method for the interpretable modelling, learning, prediction, and control of complex human behaviours, which is realised by differentiable programming.
  Solely from input features and choice outcomes without any prior knowledge, Diff-DCM can estimate interpretable closed-form utility functions that reproduce observed behaviours.
  Comprehensive experiments with both synthetic and real-world data demonstrate that Diff-DCM can be applied to various types of data and requires only a small amount of computational resources for the estimations, which can be completed within tens of seconds on a laptop without any accelerators.
  In these experiments, we also demonstrate that, using its differentiability, Diff-DCM can provide useful insights into human behaviours, such as an optimal intervention path for effective behavioural changes.
  This study provides a strong basis for the fully automated and reliable modelling, prediction, and control of human behaviours.
\end{abstract}


\section{Introduction}
Discrete choice modelling (DCM), exemplified by the well-known multinomial logit model~\cite{mcfadden_1973}, depicts the choice of an individual facing a finite set of choice alternatives based on random utility theory~\cite{thurstone_1927}\footnote[0]{This work has been published in {\it IEEE Access}. The final version is available~\cite{makinoshima_2025}.}
Many decision-making processes or behaviours can be represented as choice; hence, DCM has broad applications.
For example, human mobilities can be explained largely by combinations of travel mode choice, destination choice, and route choice.
Because of its versatility, DCM has been utilised in various fields, such as transportation, health, environmental, and consumer studies over the last fifty years~\cite{haghani_2021}.
DCM also has good compatibility with agent-based modelling and has been popularly used to model the decision-making of agents in agent-based simulations.
To date, various behaviours in a variety of agent-based models have been modelled with DCM, such as decision-making in agent-based models of migration~\cite{klabunde_2016}, decision-making of suppliers in market simulations~\cite{holm_2016}, travel mode choice in traffic simulations~\cite{horl_2018}, and dynamics modelling or exit and route choice in crowd simulations~\cite{antonini_2006, haghani_2019, nishida_2023}.

In conventional DCM, simple linear-in-parameter utility functions have often been used~\cite{train_2003}.
However, although linear utility functions have advantages in terms of explainability and ease of parameter estimation, simple utility functions cannot fully express the complex choice tendencies that occur in the real world.
Additionally, the specifications of these models, i.e.~the choice of independent variables and the form of utility functions, have depended heavily on the knowledge of the modellers.
The requirement of expert knowledge in model specification not only results in tremendous modelling cost for the design but also induces bias in the model, which may cause a mismatch between the model and real human behaviour.

To address this problem, recent studies have attempted to utilise machine learning for choice modelling~\cite{vancranenburgh_2022}.
A straightforward application of machine learning to enhance DCM is to express nonlinear utility functions using neural nets~\cite{wang_2020}.
Although this approach enables a data-driven construction of complex utility functions, the resulting models are black-box in nature and lack sufficient reliability.
Reliability becomes critical when the models are used to predict human behaviour for policy making and intervention planning.
To increase interpretability and reliability, recent studies~\cite{sifringer_2020,han_2022} proposed grey-box modelling approaches, in which neural nets are used as parts of interpretable hand-crafted utility functions.
These approaches can increase the reliability of the models while ensuring the ability to express complex utility functions; however, the models still require the specification of the utility functions to be done by modellers with domain-specific expert knowledge.
Consequently, the fully automated but interpretable modelling of complex human behaviours has remained a long-standing challenge, which we address in this study.

Attempts at automated behavioural modelling from observations have recently been made in the field of inverse generative social science (IGSS)~\cite{epstein_2023}, in which agent architectures are produced as outputs but not as inputs~\cite{vu_2019}.
A recent study~\cite{greig_2023} reported that interpretable behavioural models of agents can be estimated from behavioural observations with domain-independent primitives.
IGSS usually relies on genetic programming, which generally requires many trials of model executions for the estimations and thus tends to be computationally expensive.
Because various combinations of input variables can be considered in constructing utility functions in DCM, the search space of the utility functions tends to become excessively large, making it computationally prohibitive to estimate the utility functions with genetic programming.
Therefore, computationally more efficient techniques are required for automated identification of utility functions in DCM.

Recently, a new simulation modelling paradigm utilising differentiable programming, i.e.~differentiable simulation, is emerging, which is useful for addressing the problem in this study.
In differentiable simulations, the models are implemented to be differentiable, such that gradients obtained via automatic differentiation~\cite{baydin_2017} can be used for efficient learning and control.
Thus far, differentiable simulation has been intensively applied to physical problems~\cite{filipe_2018,degrave_2019,hu_2019}, but the scope of its application has recently been extending to agent-based social simulations~\cite{andelfinger_2022,chopra_2023,arnau_2023,joel_2023}.
Recent studies have reported that the differentiability of the simulator enables system identification~\cite{jatavallabhula_2021} or parameter estimation~\cite{chopra_2023} based on the loss between simulation and observation.

In this paper, inspired by these differentiable simulations, we develop the differentiable discrete choice model (Diff-DCM), which enables fully data-driven but interpretable behavioural modelling of complex human behaviours.
Utilising automatic differentiation enabled by differentiable programming, Diff-DCM can efficiently estimate interpretable utility functions solely from input features and choice outcomes.
The key idea behind Diff-DCM is to make the estimation of the interpretable behavioural model equivalent to a learning process of a neural net with a specific architecture.
As a result, the model estimations are performed using efficient gradient-based optimisations, which can be completed within tens of seconds even on a laptop without any accelerators.
We also demonstrate that the differentiability of Diff-DCM further enables optimal intervention planning for behavioural changes and sensitivity analysis allowing for a better understanding of human behaviours.
Although Diff-DCM is only the first step, it paves the way towards the automated modelling, prediction, and control of human behaviours for an improved society.

The remainder of this paper is organised as follows:
Section 2 explains the methods.
Section 3 describes the experimental setup to evaluate Diff-DCM.
Section 4 reports the experimental results.
Section 5 presents discussion to conclude this paper.

\section{Methods}

\subsection{Diff-DCM Architecture}
\begin{figure}[H]
  \centering
  \includegraphics[width=0.7\linewidth]{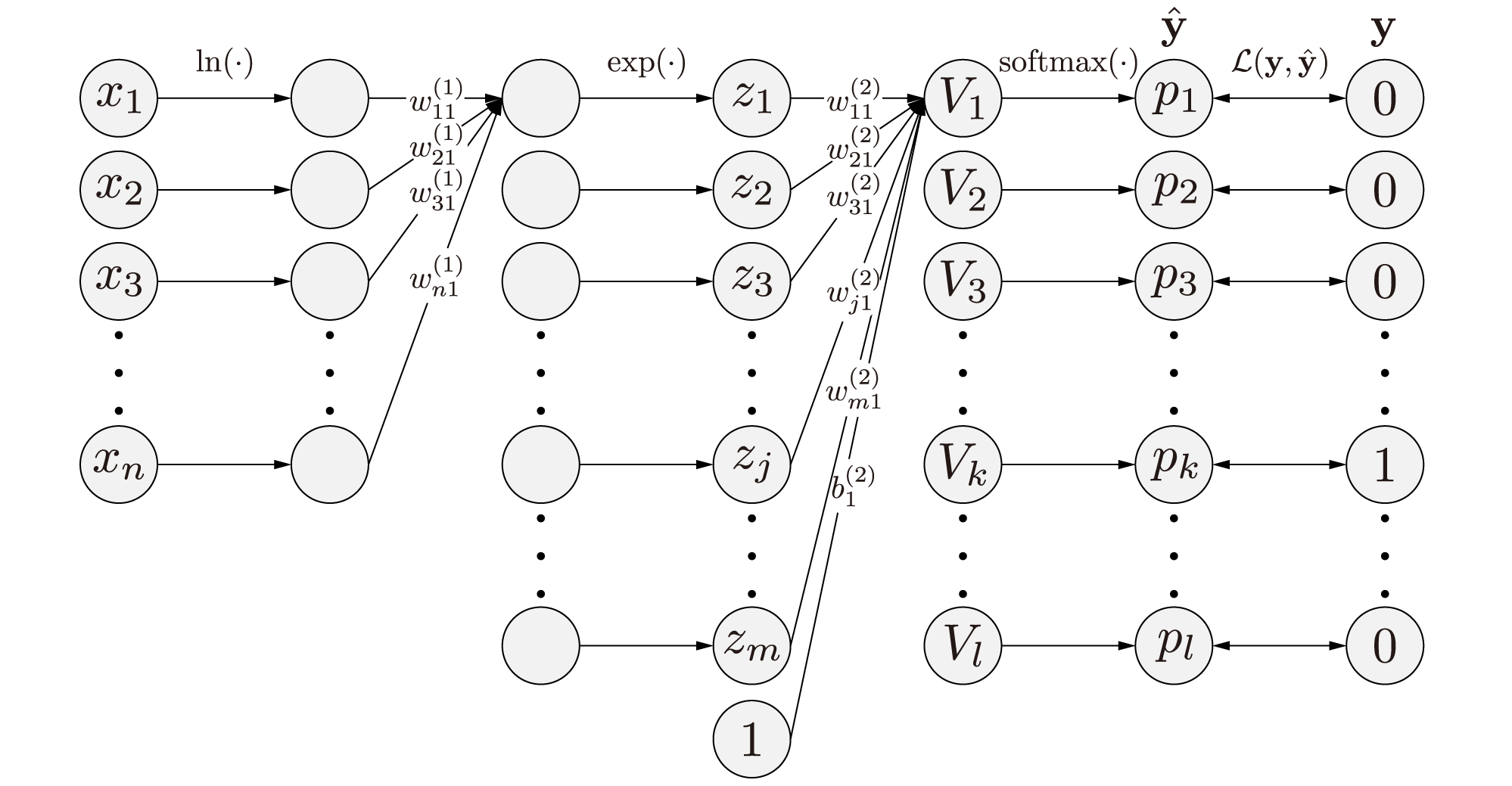}
  \caption{Architecture of Diff-DCM.}
  \label{fig:diff-dcm}
\end{figure}

Figure~\ref{fig:diff-dcm} illustrates the architecture of Diff-DCM.
We first apply the $\rm{ln}(\cdot)$ function to the input variables $x_i$ before the calculation in the subsequent first fully connected (FC) layer without bias.
Subsequently, because of the mathematical properties of the log function, the weight from the $i$-th node to the $j$-th node of the intermediate layer, $w^{(1)}_{ij}$, becomes the exponent of the variable $x_i$.
Additionally, the log function turns additions into multiplications in the first FC layer calculations.
As a result, after applying exponential transformation $\rm{exp}(\cdot)$, which is the inverse of log transformation, we have the following value $z_j$ in a node of the intermediate layer:
\begin{eqnarray}
  z_j = \prod_{i=1}^{n}x_{i}^{w^{(1)}_{ij}}.
\end{eqnarray}
We find that $z_j$ can become a variety of terms such as higher-order or interaction terms of input variables $x_i$, depending on the value of the weight $w^{(1)}_{ij}$.

Further proceeding with the calculation in the second FC layer with a bias, we obtain the following utility function $V_k$:
\begin{eqnarray}
  \label{eq:utility}
  V_k = \sum_{j=1}^{m} w^{(2)}_{jk} \prod_{i=1}^{n}x_{i}^{w^{(1)}_{ij}} + b^{(2)}_k,
\end{eqnarray}
where $w^{(2)}_{jk}$ and $b^{(2)}_k$ are the weights and biases of the second FC layer, respectively.
From Eq.~(\ref{eq:utility}), we can find that the utility functions here have the form of polynomials in input variables $x_i$ but with real-number exponents.

By applying the $\rm{softmax}(\cdot)$ function to the utility functions, we can convert the utilities into choice probabilities $p_k$, enabling the computation of the loss between the ground truth and the predicted choice, i.e.~$\mathcal{L}(\mathbf{y}, \hat{\mathbf{y}})$.
For the loss function, we use the cross-entropy loss, given as
\begin{eqnarray}
  \mathcal{L}(\mathbf{y}, \hat{\mathbf{y}}) = - \sum_{k=1}^{l}y_k {\rm{log}}(p_k),
\end{eqnarray}
where $y_k$ denotes the one-hot encoded true choice, and $p_k$ is the predicted probability.
Minimising this loss function is equivalent to maximising the log-likelihood function for parameter estimation in conventional DCM~\cite{sifringer_2020}.
Therefore, Diff-DCM is the differentiable implementation of DCM but with more complex utility functions than in conventional modelling with simple linear utility functions.
By minimising this loss function, we can obtain the parameters $w^{(1)}_{ij}$, $w^{(2)}_{jk}$, and $b^{(2)}_k$ that reproduce the given choice outcomes.

After the training, because Diff-DCM has an architecture in which the weights ($w^{(1)}_{ij}$ and $w^{(2)}_{jk}$) and bias ($b^{(2)}_k$) correspond to the coefficients, exponents, and alternative-specific constants of the utility functions, we can extract interpretable closed-form utility functions from the trained model.
Consequently, we can automatically estimate the utility functions solely from the input variable $\mathbf{x}$ and choice observation $\mathbf{y}$.
Because this estimation is executed as the learning process of a neural net with gradient-based optimisation, the estimation is conducted in an efficient way and is thus fast.

In this study, we used PyTorch~\cite{paszke_2019} to implement Diff-DCM.
Listing~\ref{ddcmcode} shows Diff-DCM implemented using PyTorch.
As shown in the snippet, the architecture of Diff-DCM is simple and easy to implement.
Because of the use of the $\rm{ln}(\cdot)$ function on the inputs, the zero or negative inputs need some prior transformation to prevent the occurrence of NaN values.
In this study, we applied normalisation to the data and added a small value (e.g.~$\epsilon=1e^{-4}$) to the zero values.
The number of nodes in the intermediate layer is a parameter that corresponds to the number of terms of the utility functions, i.e.~expressiveness.
We can arbitrarily set the number of nodes such that the size becomes sufficient for expressing the utility functions.
This study considered 10 and 24 nodes for synthetic and real-world datasets, respectively; these are considered sufficiently large based on the number of input variables.

\begin{figure}[H]
  \begin{lstlisting}[language=Python, caption=Diff-DCM implemented with PyTorch, label=ddcmcode]
    class diff_dcm(nn.Module):  
        def __init__(self):
            super().__init__()
            self.fc1 = nn.Linear(n_features, n_nodes, bias=False)
            self.fc2 = nn.Linear(n_nodes, n_choice, bias=True)
    
        def forward(self, x):
            x = torch.log(x)
            x = self.fc1(x)
            x = torch.exp(x)
            x = self.fc2(x)
            return x
\end{lstlisting}
\end{figure}

\subsection{Training}
To train Diff-DCM, an Adam optimiser~\cite{kingma_2017} with a learning rate of $\alpha=1e^{-3}$ was used.
We trained the model for 100 epochs with a batch size of 50.
For the problems considered in this study, we observed that 100 epochs were sufficient for observing the loss convergence.
These training parameters were consistently used throughout this study.

Although the utility functions can be obtained as closed-form expressions, it might be difficult to comprehend what are meant by the terms with real-number exponents.
Furthermore, it is desirable to prevent the utility functions from becoming exceedingly complex for better explainability.
In such cases, we can utilise regularisation and fine tuning to obtain simplified utility functions.
In this study, we trained the model with a weight decay of $1e^{-2}$, where a more simplified function form is desirable.
In addition to the initial training, we also conducted fine tuning to make terms with integer exponents.
After the training, we applied {\lstinline|torch.round|} function to the weights in the first FC layer $w^{(1)}_{ij}$, yielding utility functions with integer exponents.
Then, we fixed the weight of the first FC layer $w^{(1)}_{ij}$ and re-trained the model parameters $w^{(2)}_{jk}$ and $b^{(2)}_k$ of the second FC layer.
These processes simplify the form of the utility functions, leading to better explainability.
In this study, we used the weight decay for the synthetic data but not for the real-world data so as to avoid oversimplifying the utility functions.
Because real-world behaviours are often complex, these simplifications may not be suited for modelling complex behaviours and may reduce modelling capability.
Therefore, this simplification process should be considered as optional for real-world problems.

\subsection{Analyses with Gradients}
\subsubsection{Sensitivity Analysis}
Gradients available from automatic differentiation can be utilised for various analysis~\cite{arnau_2023}.
The first use case of the gradients in this study is sensitivity analysis.
Although the utility functions are estimated as closed-form expressions, the function can be nonlinear, i.e.~a combination of higher-order terms and interaction terms.
As a result, the impact of a certain variable $x_i$ on a utility, i.e.~sensitivity, is not obvious at a glance.
In this case, once we run Diff-DCM and obtain the computational graph, the sensitivity is given by $\frac{\partial V}{\partial x}$, which is easily obtained via automatic differentiation.
This sensitivity is equivalent to what is referred to as marginal utility in conventional DCM and in the field of economics~\cite{train_2003}.
If the utility functions are nonlinear, the derivative is not constant and changes depending on the individual.
In such cases, we can calculate the average marginal utility over all $N$ individual samples, i.e.~$\frac{1}{N} \sum \frac{\partial V}{\partial x}$, to evaluate the sensitivity.

\subsubsection{Optimal Intervention Path}
The second use case of the gradients is the calculation of the optimal intervention path to cause behavioural changes.
In practical applications, there are some cases wherein we want to cause behavioural changes in people.
For instance, one may want to promote pro-environmental behaviours for environmental conservation or vaccinations to reduce the spread of infectious disease, either of which can be modelled with DCM~\cite{arnau_2023,paszke_2019}.
When planning an intervention aimed at inducing behavioural change, policy makers would find it essential to know the optimal intervention path to realising the desired behavioural change.
This intervention path can be calculated using automatic differentiation.
Given the desired choice $\mathbf{y}^*$, an individual with features $\mathbf{x}$, and the trained Diff-DCM $\phi$, the optimal direction to making the choice of the individual $\phi(\mathbf{x})$ close to the desired choice $\mathbf{y}^{*}$ is given by the gradient of the loss with respect to $\mathbf{x}$.
Therefore, the path to inducing behavioural change can be obtained by repeatedly updating $\mathbf{x}$ via gradient descent:
\begin{eqnarray}
  \mathbf{x}^{(t+1)} = \mathbf{x}^{(t)} - {\gamma} \frac{\partial \mathcal{L}(\mathbf{y}^{*}, \phi(\mathbf{x}))}{\partial \mathbf{x}},
\end{eqnarray}
where $t$ denotes the step of updates, and $\gamma$ is a step size.
In the demonstration presented later in this paper, we obtained the path by updating $\mathbf{x}$ for 100 steps with a step size of $\gamma = 0.05$.

\section{Experimental Setup}
\subsection{Experiments with Synthetic Data}
\subsubsection{Data from Simple Linear Utility}
We first generated synthetic data with simple linear utility functions as the most fundamental case.
For the simple linear utility functions, the following utility functions were assumed for three alternatives:
\begin{align}
  \begin{split}
     & V_1 = 2.0x_1 - 1.0x_2,       \\
     & V_2 = -1.0x_1 + 2.0x_2,      \\
     & V_3 = 1.0x_1 + 1.0x_2 - 2.0,
  \end{split}
\end{align}
where $x_1$ and $x_2$ are feature variables sampled from uniform distributions, i.e.~$x_1 \sim \mathcal{U}(0, 10)$ and $x_2 \sim \mathcal{U}(0, 10)$.
The choice probability is calculated using the softmax function and is used in the {\lstinline|numpy.random.choice|} function to synthesise the choice outcomes.
We prepared 10000 samples for training and 1000 for testing.

\subsubsection{Data from Utility with a Dummy Variable}
Because feature variables for choice modelling often include qualitative variables, we synthesised the second dataset using the following utility functions with a dummy variable for three alternatives:
\begin{align}
  \begin{split}
     & V_1 = 3.0x_1 - 1.0x_2,          \\
     & V_2 = -1.0x_1 + 3.0x_2,         \\
     & V_3 = 1.5x_1 + 1.5x_2 + 3.0x_3.
  \end{split}
\end{align}
Here, whereas $x_1$ and $x_2$ are quantitative feature variables sampled from uniform distributions, i.e.~$x_1 \sim \mathcal{U}(0, 10)$ and $x_2 \sim \mathcal{U}(0, 10)$, $x_3$ is a qualitative variable, which was sampled with $\mathrm{Bernoulli}(0.7)$ and takes a value of either 0 or 10.
With the same procedure for the synthesis of the first dataset, 10000 and 1000 samples were generated for training and testing, respectively.

\subsubsection{Data from Nonlinear Utility}
For a more complex case, we synthesised a third dataset with the following nonlinear utility functions for three alternatives:
\begin{align}
  \begin{split}
     & V_1 = 2.0x_1^2 - 1.0x_1x_2 - 1.0x_2^2,       \\
     & V_2 = -1.0x_1^2 - 1.0x_1x_2 + 2.0x_2^2,      \\
     & V_3 = -0.5x_1^2 + 2.5x_1x_2 -0.5x_2^2 - 1.0,
  \end{split}
\end{align}
where $x_1 \sim \mathcal{U}(0, 10)$ and $x_2 \sim \mathcal{U}(0, 10)$.
The functions have both higher-order terms and interaction terms, which cannot be expressed via conventional DCM with linear utility functions.
With the same procedure as that used for the previous data synthesis, we synthesised 10000 and 1000 samples for training and testing, respectively.

\subsubsection{Data from Logical Conditions}
The data generation mechanisms behind real-world choice outcomes are usually unknown; therefore, mismatches between observations and models~\cite{gaskin_2023,makinoshima_2024} are likely to occur in practical applications.
Assuming such a challenging situation, a fourth dataset with four choice alternatives was generated not by utility functions but by logical conditions (Algorithm~\ref{algo}).
Although this algorithm is simple, the synthesised data are difficult to represent with utility functions because of the data-model mismatch and the discontinuous choice probability.
With uniform distributions, 10000 and 1000 samples were synthesised for training and test evaluation, respectively.
\begin{algorithm}[H]
  \caption{Logical conditions for data synthesis}
  \begin{algorithmic}[1]
    \STATE $x_1 \sim \mathcal{U}(0, 10)$
    \STATE $x_2 \sim \mathcal{U}(0, 10)$
    \IF{$x_1 \ge 5.0 $ and $x_2 \ge 5.0$}
    \STATE choice = 1
    \ELSIF{$x_1 < 5.0$ and $x_2 \ge 5.0$}
    \STATE choice = 2
    \ELSIF{$x_1 < 5.0$ and $x_2 < 5.0$}
    \STATE choice = 3
    \ELSIF{$x_1 \ge 5.0$ and $x_2 < 5.0$}
    \STATE choice = 4
    \ENDIF
  \end{algorithmic}
  \label{algo}
\end{algorithm}
\subsubsection{Evaluation}
For the synthetic datasets generated with utility functions, we can know the true function forms and compare the estimated utility functions directly with the true functions.
For our evaluation of the estimation performance, we confirmed the generated terms and the directions of coefficients of the estimated utility functions.
Because the choice probability is determined only by the differences among utilities of choice alternatives in DCM~\cite{train_2003}, we examined the directions of the coefficients rather than the absolute values and normalised the utility values for comparison.
For the dataset synthesised with logical conditions, we verified the estimated choice probability distributions against the true distributions.
Additionally, we verified the forecasting performance of the estimated models by investigating the accuracies of the predicted choices against the ground truth choice outcomes in the test datasets.

\subsection{Experiments with Real-world Data}
\subsubsection{Data Description}
For the real-world data, we used the Swissmetro dataset~\cite{bierlaire_2001}, which provides information regarding the preferences of respondents relevant to evaluating the impact of a new transportation mode, Swissmetro (SM), against conventional transportation modes (car and train).
We employed this dataset because it has been often used in past DCM research studies and is publicly available as an example dataset for an open-source DCM software, BIOGEME~\cite{bierlaire_2003, bierlaire_2023}.
For details on the dataset such as the data collection procedures, the readers are referred to the original paper~\cite{bierlaire_2001} and an additional document~\cite{bierlaire_2018}.
As in the preparation of the synthetic dataset, the data were normalised between 0 and 10 and used for training and testing.
We screened this dataset following the data screen procedure applied in an earlier study~\cite{sifringer_2020} on the same dataset and yielded a total of 9036 observations.
These 9036 observations were then divided into 7236 training and 1800 test samples, corresponding to approximately 80\% and 20\% of the total number of samples, respectively.

\subsubsection{Evaluation}
For real-world data, we cannot know the true mechanism by which the data were generated; therefore, we compared Diff-DCM with a benchmark DCM with the following expert-designed utility functions~\cite{bierlaire_2001}:
\begin{align}
  \begin{split}
    V_{train} = {} & \beta_{T}T_{train} + \beta_{C}C_{train} + \beta_{Freq}Freq_{train} \\
                   & + \beta_{GA}GA + \beta_{Age}Age, \nonumber
  \end{split}                \\
  \begin{split}
    V_{SM} = {} & ASC_{SM} + \beta_{T}T_{SM} + \beta_{C}C_{SM} + \beta_{Freq}Freq_{SM} \\
                & + \beta_{GA}GA + \beta_{Seats}Seats,
  \end{split}                 \\
  V_{car} = {} & ASC_{car} + \beta_{T}T_{car} + \beta_{C}C_{car} + \beta_{Luggage}Luggage, \nonumber
\end{align}
where $ASC$ are the alternative-specific constants normalised to that of train ($ASC_{train} = 0$), $T$ denotes travel time, $C$ denotes travel cost, $Freq$ denotes headway, $GA$ is a dummy variable indicating whether an individual has the annual pass, $Age$ represents age in five classes, $Luggage$ refers to pieces of luggage, $Seats$ is a dummy variable representing seats configuration, and $\beta$ are parameters for the aforementioned variables to be estimated.
This benchmark model was estimated using BIOGEME~\cite{bierlaire_2003, bierlaire_2023}, in which the parameters $\beta$ were estimated using Newton/BFGS with trust region for simple bounds~\cite{conn_1988}.
For Diff-DCM and the benchmark model, the same training and test samples were used in model estimation and prediction accuracy evaluation.
For the evaluation, we verified whether Diff-DCM could provide useful insights equivalent to those from the expert-designed benchmark model, even without any prior knowledge or assumptions.
With the test dataset, the choice prediction performance was also evaluated in terms of the prediction accuracy and the log-likelihood value.

\subsection{Computational Setup}
All experiments were conducted on a laptop with 11th Gen Intel\textsuperscript{\textregistered} Core\textsuperscript{\texttrademark} i7-1185G7~@~3.00~GHz with 16~GB memory.
We did not use any GPUs in our experiments because the estimations and predictions by Diff-DCM were sufficiently fast, even without GPUs.

\section{Results}
\subsection{Performance on Synthetic Data}
\subsubsection{Simple Linear Utility}
\begin{figure}[b]
  \centering
  \includegraphics[width=0.7\linewidth]{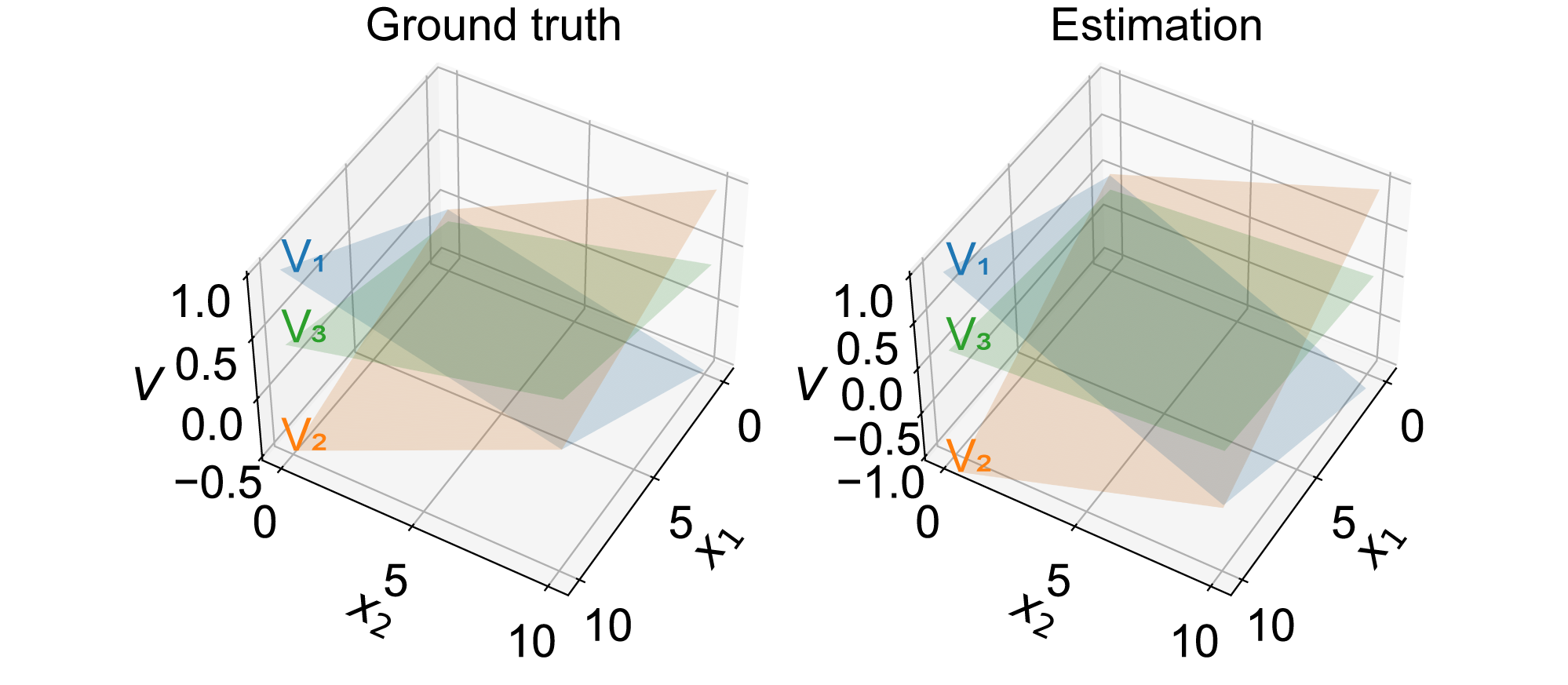}
  \caption{Estimation result for linear utility functions.}
  \label{fig:est_linear}
\end{figure}

For the data synthesised from simple linear utility functions, Diff-DCM estimated the following utility functions, which are visualised with ground truth utility functions in Fig.~\ref{fig:est_linear}:
\begin{align}
  \begin{split}
     & V_1 = 1.037x_1 - 1.403x_2 + 0.520,  \\
     & V_2 = -1.466x_1 + 1.046x_2 + 0.702, \\
     & V_3 = 0.276x_1 + 0.202x_2 - 1.222.
  \end{split}
\end{align}
No prior assumption on function forms was applied in the estimation; nevertheless, Diff-DCM successfully recovered the linear utility functions solely from data.
Although the magnitudes of the coefficients in $V_1$ and $V_2$ were slightly different from the ground truth, major characteristics such as generated terms and directions of the coefficients were successfully estimated, as can be seen in Fig.~\ref{fig:est_linear}.
Consequently, the estimated model achieved an accuracy of 89.0\% in its prediction on the test dataset, which was equivalent to the performance of the ground truth model (89.2\%).

\subsubsection{Utility with a Dummy Variable}
\begin{figure}[t]
  \centering
  \includegraphics[width=0.7\linewidth]{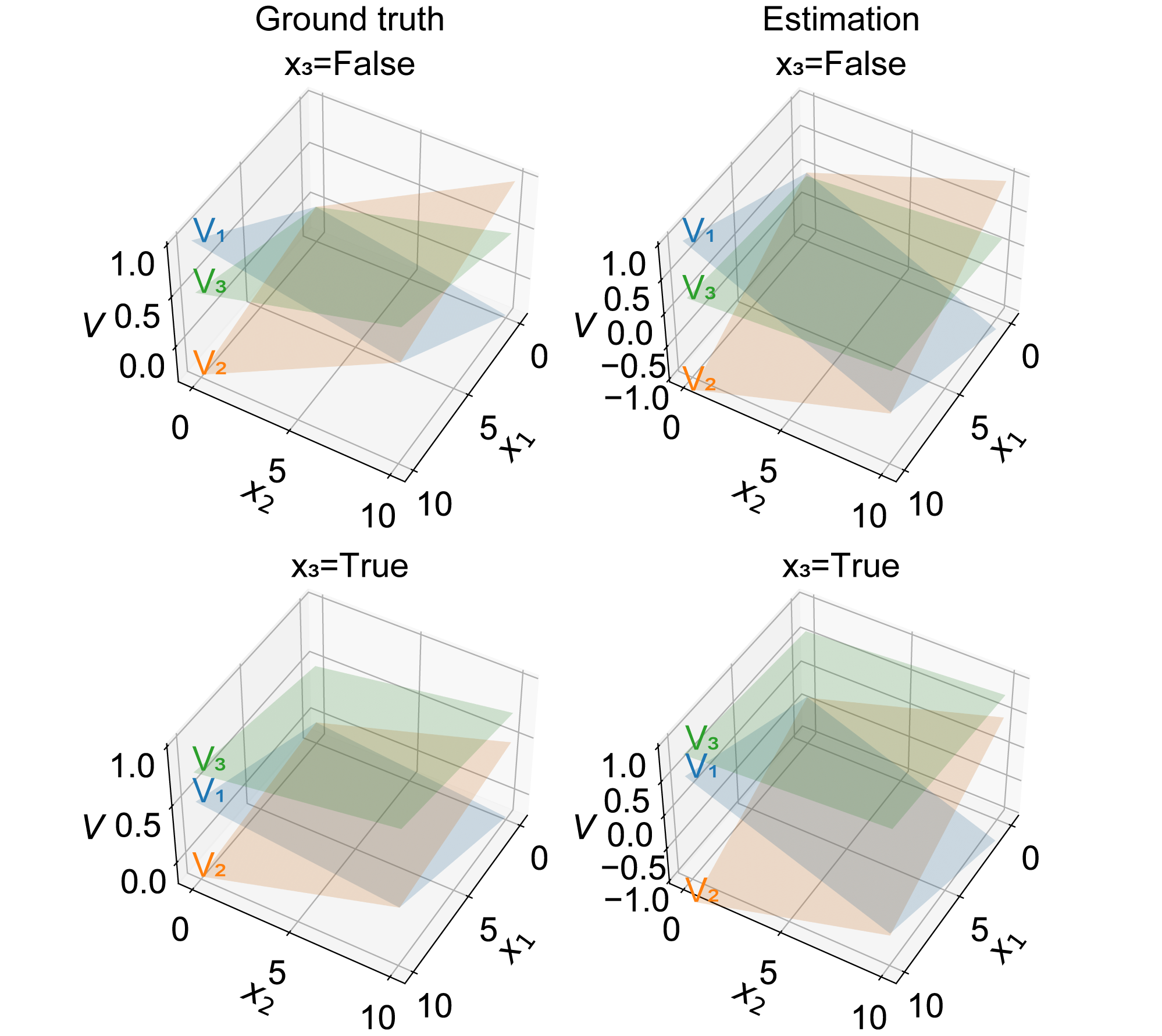}
  \caption{Estimation result for utility functions with dummy variable.}
  \label{fig:est_dummy}
\end{figure}

Diff-DCM estimated the following utility functions for data generated from the utility functions with a dummy variable:
\begin{align}
  \begin{split}
     & V_1 = 0.855x_1 - 1.077x_2 -0.340x_3 + 0.126,   \\
     & V_2 = -1.098x_1 + 0.855x_2 - 0.355x_3 + 0.160, \\
     & V_3 = 0.176x_1 + 0.179x_2 + 0.694x_3 - 0.291.
  \end{split}
\end{align}
Together with the ground truth utility functions, these estimated utility functions are visualised in Fig.~\ref{fig:est_dummy}.
Similar to the result in the case of the linear utility functions, Diff-DCM successfully recovered the linear utility functions with the dummy variable solely from data.
To reproduce the ground truth functions, which have $x_3$ increasing the utility $V_3$ when true, Diff-DCM estimated $V_1$ and $V_2$ having $x_3$ with negative coefficients and $V_3$ with positive $x_3$.
When $x_3$ is true, these terms relatively increase $V_3$, which has the same effect as that of $x_3$ in the ground truth utility function $V_3$.
In this way, Diff-DCM successfully reproduced the choice outcomes, which resulted in an accuracy of 96.4\% in its prediction on the test data samples.
This prediction accuracy was equivalent to the performance of the ground truth utility functions (96.5\%).

\subsubsection{Nonlinear Utility}
\begin{figure}[t]
  \centering
  \includegraphics[width=0.7\linewidth]{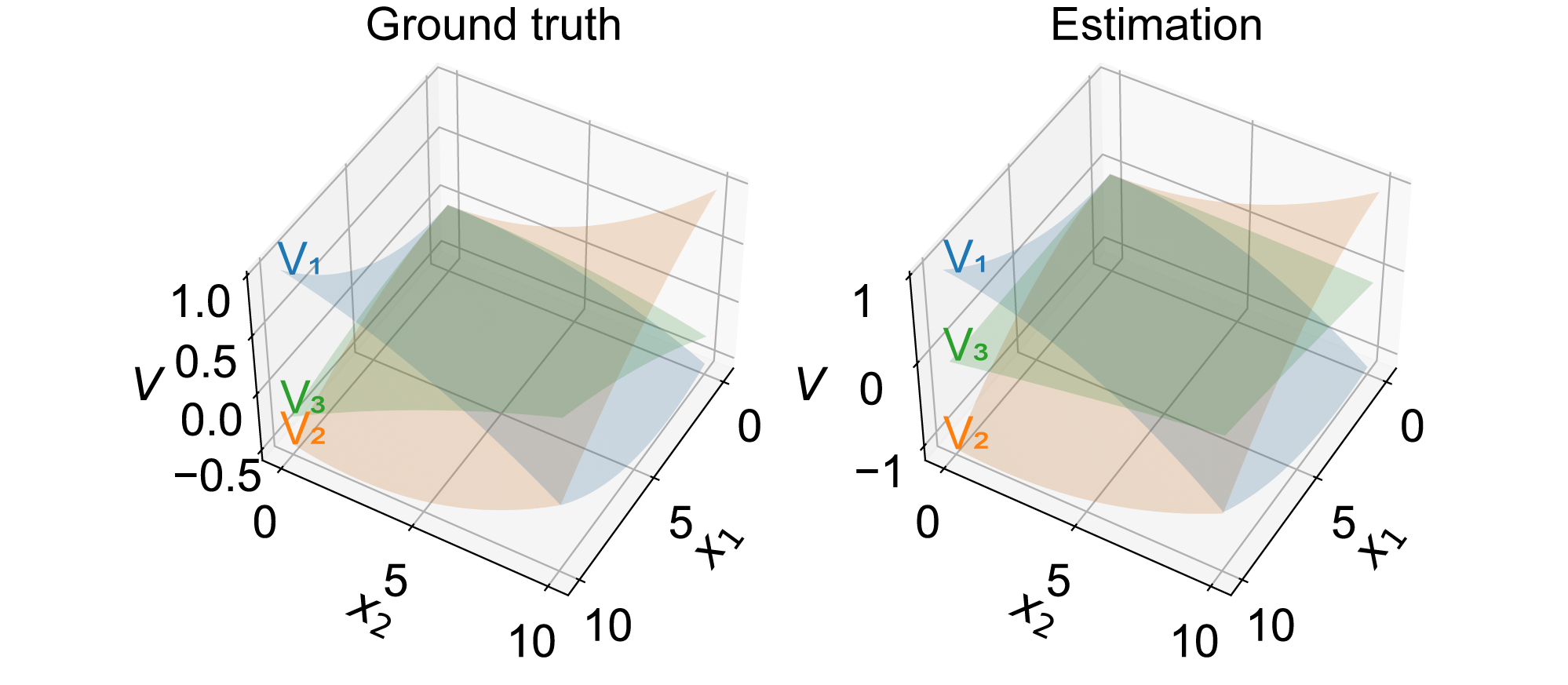}
  \caption{Estimation result for nonlinear utility functions.}
  \label{fig:est_nonlinear}
\end{figure}

Diff-DCM recovered the following utility functions from the data synthesised by nonlinear utility functions:

\begin{align}
  \begin{split}
     & V_1 = 0.778x_1^2 - 0.279x_1x_2 - 0.853x_2^2 + 0.255,  \\
     & V_2 = -0.863x_1^2 - 0.260x_1x_2 + 0.760x_2^2 + 0.142, \\
     & V_3 = -0.036x_1^2 + 0.438x_1x_2 - 0.047x_2^2 - 0.404.
  \end{split}
\end{align}
Although the ground truth utility functions have complex forms that include higher-order and interaction terms, Diff-DCM successfully estimated similar utility functions without any prior assumptions, as shown in Fig.~\ref{fig:est_nonlinear}.
The generated terms and the directions of the coefficients were equivalent to those of the ground truth functions, i.e.~the estimated utility functions reproduced the major features of the ground truth functions.
As a result, the estimated functions predicted the choice outcomes in the test data with an accuracy of 97.1\%, as accurately as did the ground truth model (97.5\%).
These results verified that, regardless of the complexity and kinds of variables involved, Diff-DCM can recover utility functions solely from input features and choice outcomes.

\subsubsection{Logical Conditions}
\begin{figure}[t]
  \centering
  \includegraphics[width=0.7\linewidth]{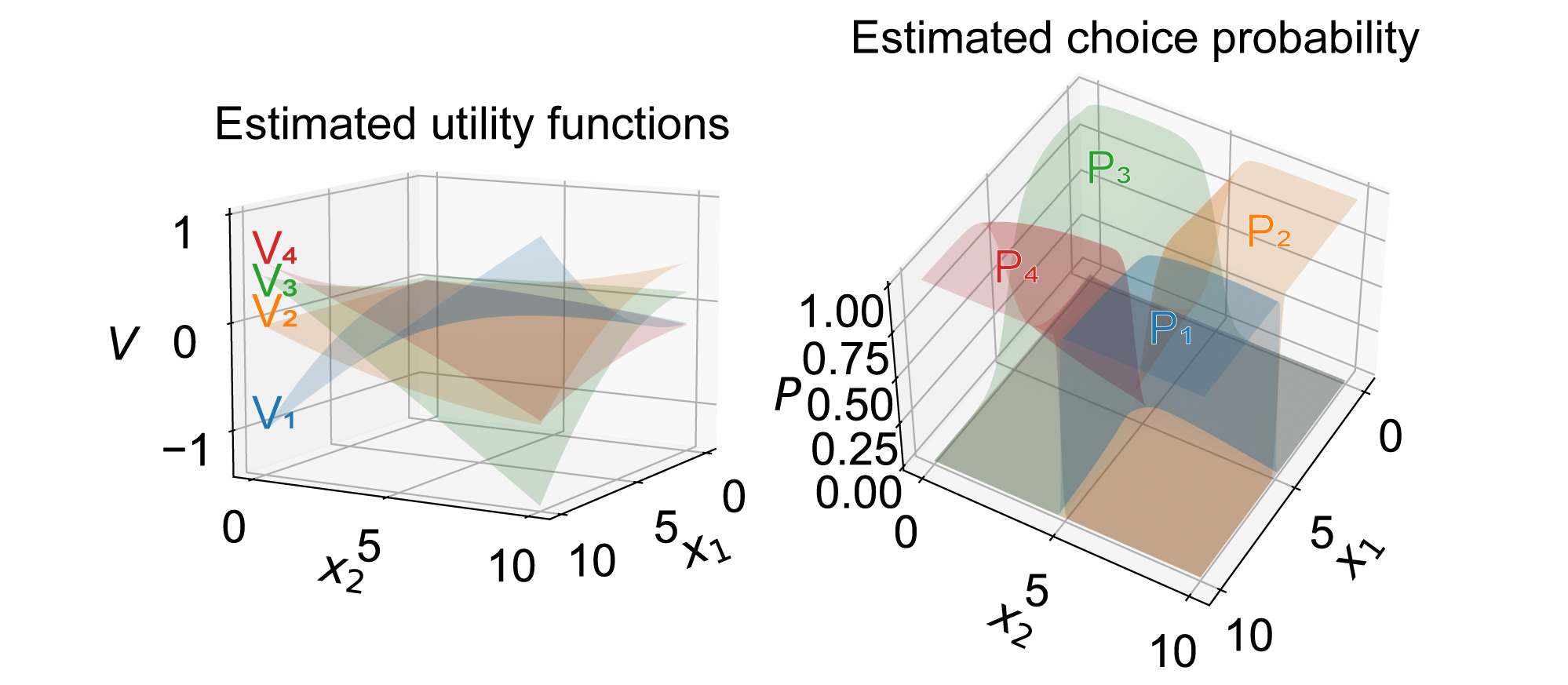}
  \caption{Estimation results for utility functions reproducing logical conditions.}
  \label{fig:est_logical}
\end{figure}

The following complex utility functions were estimated from the data synthesised with logical conditions:

\begin{align}
  \begin{split}
     & V_1 = 0.231x_1^2x_2 - 0.114x_2^2 - 1.051x_1^2 - 0.855x_2 - 0.689,  \\
     & V_2 = -0.107x_1^2x_2 + 0.432x_2^2 - 0.028x_1^2 - 0.192x_2 - 0.855, \\
     & V_3 = -0.197x_1^2x_2 - 0.077x_2^2 + 0.413x_1^2 + 1.557x_2 + 3.094, \\
     & V_4 = -0.102x_1^2x_2 - 0.159x_2^2 + 0.590x_1^2 - 0.484x_2 - 1.551.
  \end{split}
\end{align}
The estimated functions and choice probabilities are visualised in Fig.~\ref{fig:est_logical}.
Here, the utility values were converted into choice probabilities using the softmax function for visualisation.
From the probability view in Fig.~\ref{fig:est_logical}, we can see that the complex nonlinear functions estimated by Diff-DCM reproduced the choice outcomes equivalent to the logical condition used for data synthesis (Algorithm~\ref{algo}).
The estimated nonlinear utility functions achieved a prediction accuracy of 99.7\% on the test dataset.
This result validated that, with complex nonlinear utility functions, Diff-DCM can reproduce the choice behaviour that is not based on utility functions in DCM, supporting the applicability of Diff-DCM to real-world data.

\subsubsection{Optimal Intervention Path}
\begin{figure}[h]
  \centering
  \includegraphics[width=0.6\linewidth]{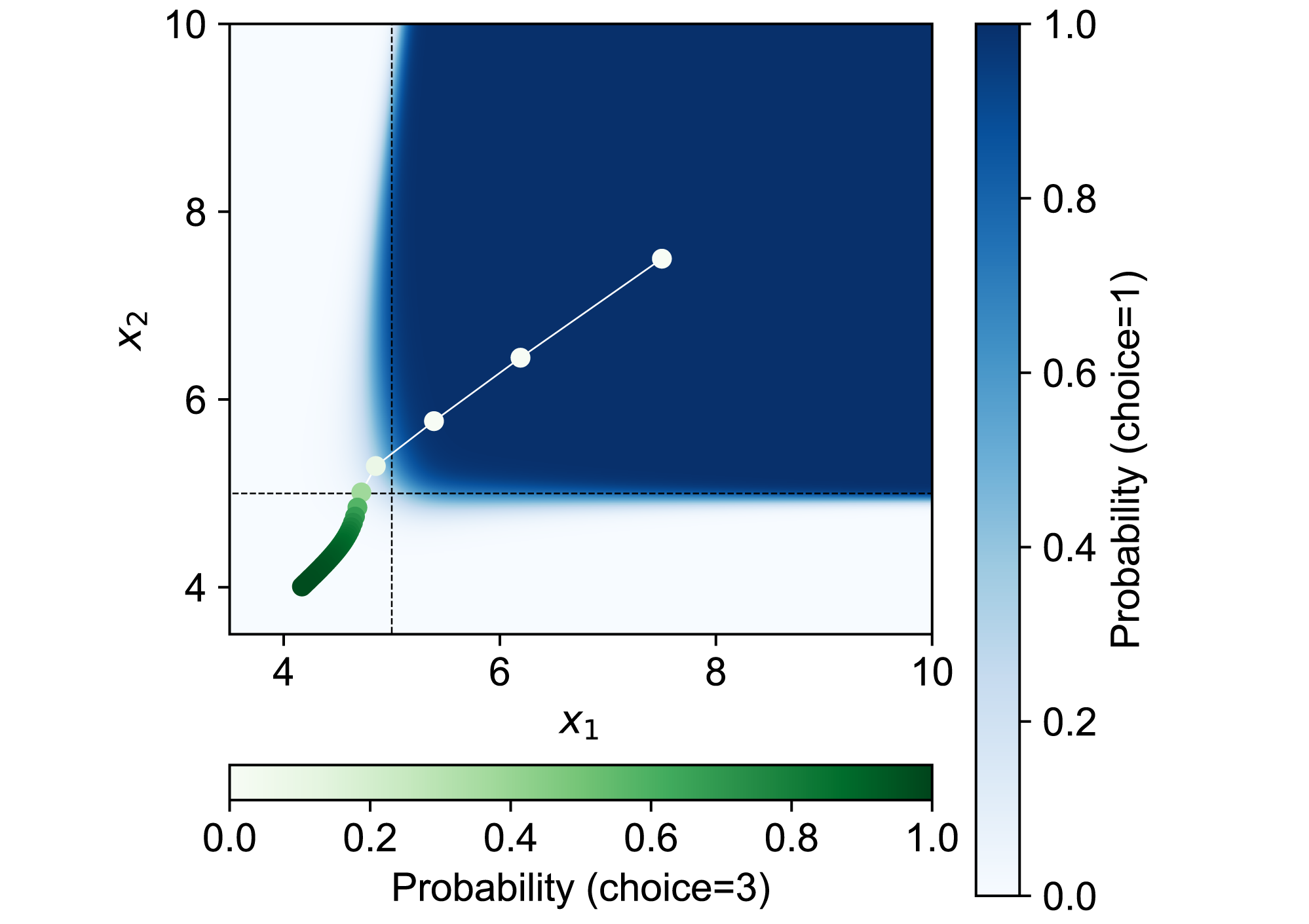}
  \caption{Optimal intervention path (from \mbox{choice 1} to \mbox{choice 3}) obtained using Diff-DCM.}
  \label{fig:optpath}
\end{figure}

Using the case of {\it logical conditions}, we demonstrate the calculation of the optimal intervention path.
Here, we considered a case in which we want to change the choice~1 of an individual with feature vector $\mathbf{x}=[7.5, 7.5]$ to choice~3.
Figure~\ref{fig:optpath} visualises the obtained optimal intervention path.
We can observe that the choice of the individual effectively changed from choice~1 to choice~3 via the obtained path.
The original logical conditions are discontinuous and thus not differentiable.
By contrast, Diff-DCM is differentiable, which enables the calculation of the optimal intervention path using gradients.

\subsubsection{Computational Cost}
\begin{figure}[t]
  \centering
  \includegraphics[width=0.7\linewidth]{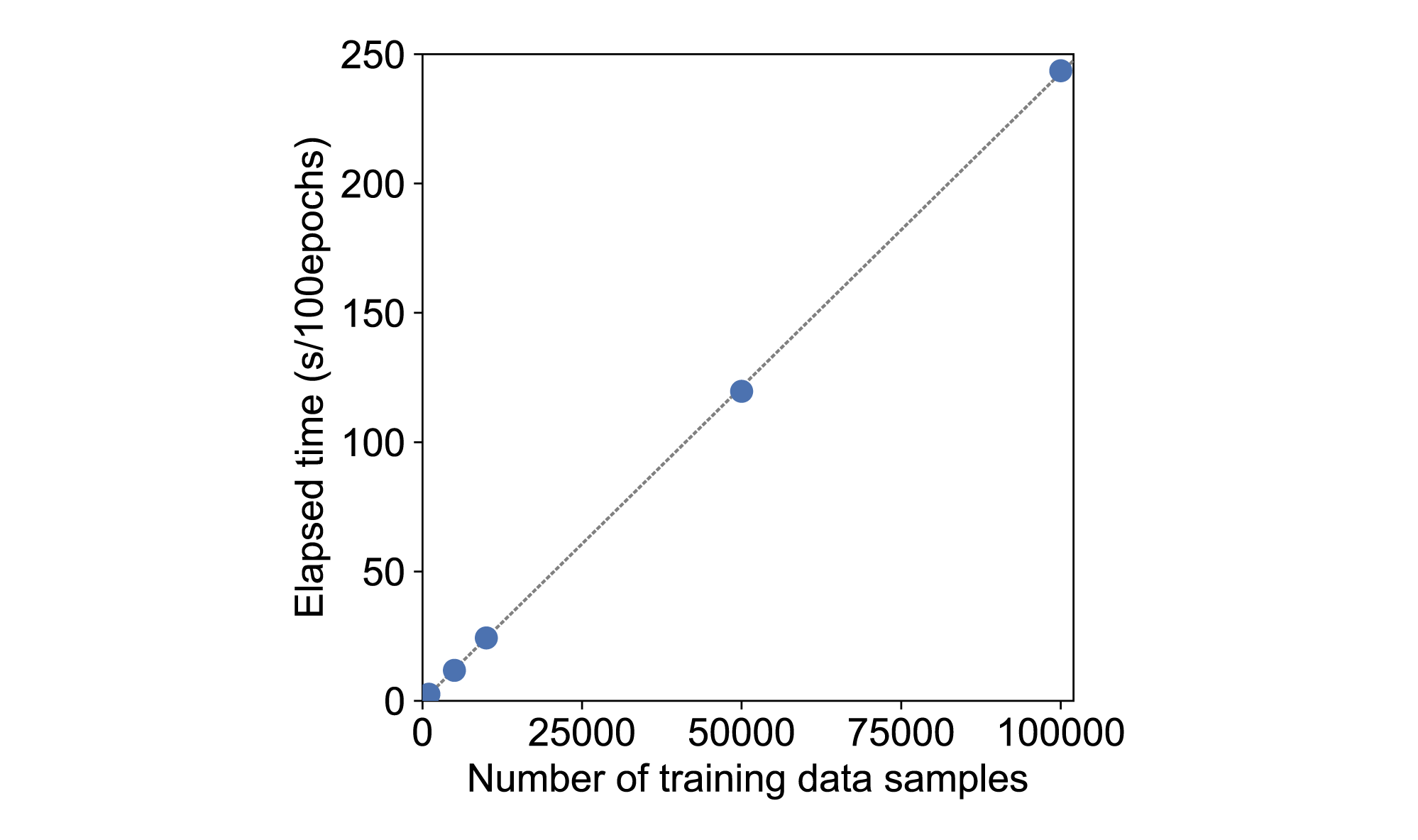}
  \caption{Scalability of Diff-DCM for larger dataset.}
  \label{fig:scalability}
\end{figure}

Figure~\ref{fig:scalability} reports the measured time required for the estimation of utility functions for different numbers of samples.
The results show that the training of Diff-DCM on 10000 samples was completed within 30~s on a laptop and scales linearly for larger data samples.
We also measured the required time for obtaining the optimal intervention path shown in Fig.~\ref{fig:optpath} and found that the calculation demonstrated in this study was completed within few seconds.
The estimation and calculation using Diff-DCM can be executed very fast even on a laptop without GPUs, supporting the practicality of Diff-DCM.

\subsection{Performance on Real-world Data}
\subsubsection{Estimation Results}
Table~\ref{tab:benchmark} presents the estimation results of the expert-designed model, which is used as a benchmark.
From the expert-designed model, we can find that the car and SM have higher utilities than that of the train by default based on the positive values for $ASC_{car}$ and $ASC_{SM}$.
We can also know that the mode choice can be explained largely by time and cost because $\beta_T$ and $\beta_C$ have large values.

Without any prior knowledge, Diff-DCM generated the utility functions consisting of the following terms: $\{ T_{car}, C_{SM}/T_{car}, 1/T_{train}, 1/(T_{train}C_{car}), T_{car}T_{SM}, 1/(C_{car}Freq_{train}Age), C_{SM}/T_{car}, T_{car}/C_{car},\allowbreak (C_{train}C_{SM})/(T_{car}T_{train}), (T_{SM}C_{train})/C_{car}, T_{car}/T_{train},\allowbreak const. \}$.
The generated terms may appear complex at first glance; however, upon examining the components, it becomes evident that the generated terms are mainly interaction terms of time and cost of travel modes.
For example, $T_{car}/C_{car}$ is the meaningful term related to the value of time~\cite{small_2012}, a well-known concept in the field of economics.
Diff-DCM automatically generated these terms without any prior knowledge about the travel mode choice.

To make the result more comprehensible, we summarised the alternative specific constants and the most influential terms estimated by Diff-DCM in Table~\ref{tab:diffdcm}.
Regarding the $ASC$, Diff-DCM estimated that the SM and car have higher utility values than that of the train by default.
This estimated relation among the travel modes is consistent with the result of the expert-designed model.
From the most influential terms estimated by Diff-DCM, we can find that relative cost and time are important factors that determine the travel mode choice, which is also consistent with the insights obtained from the expert-designed model.
These results validated that, relying on only the input variables and choice outcomes, Diff-DCM can provide useful insights equivalent to those obtained from the model designed using expert knowledge.

\begin{table}[t]
  \centering
  \caption{Estimation results of the expert-designed model.}
  \label{tab:benchmark}
  \begin{tabular}{crrrr} \toprule
    Parameters        & Estimates & Std. err. & t-stat & p-value  \\ \midrule
    $ASC_{car}$       & 0.994     & 0.185     & 5.39   & 7.15e-08 \\
    $ASC_{SM}$        & 1.84      & 0.17      & 10.8   & 0        \\
    $\beta_{Age}$     & 0.202     & 0.0513    & 3.95   & 7.91e-05 \\
    $\beta_{C}$       & -0.517    & 0.0364    & -14.2  & 0        \\
    $\beta_{GA}$      & 0.131     & 0.0166    & 7.87   & 3.55e-15 \\
    $\beta_{Freq}$    & -0.0457   & 0.00801   & -5.71  & 1.14e-08 \\
    $\beta_{Luggage}$ & -0.0259   & 0.0147    & -1.76  & 0.0777   \\
    $\beta_{Seats}$   & 0.04      & 0.011     & 3.64   & 0.000275 \\
    $\beta_{T}$       & -1.33     & 0.0918    & -14.5  & 0        \\ \bottomrule
  \end{tabular}
\end{table}

\begin{table}[t]
  \centering
  \caption{Diff-DCM estimation results. \rm{The normalised $ASC$ and coefficients for the most influential terms are shown in brackets.}}
  \label{tab:diffdcm}
  \begin{tabular}{cccc}\toprule
    \multicolumn{1}{c}{\multirow{2}{*}{Mode}} & \multicolumn{1}{c}{\multirow{2}{*}{Const. (ASC)}} & \multicolumn{2}{c}{Most influential term}                                                                               \\
    \multicolumn{1}{c}{}                      & \multicolumn{1}{c}{}                              & \multicolumn{1}{c}{Positive}                               & \multicolumn{1}{c}{Negative}                               \\ \midrule
    \multirow{2}{*}{$V_{train}$}              & \multirow{2}{*}{$-0.877~(0.0)$}                   & \multirow{2}{*}{$\frac{1}{T_{train}}~(+0.28)$}             & \multirow{2}{*}{$\frac{T_{SM}C_{train}}{C_{car}}~(-0.60)$} \\\\
    \multirow{2}{*}{$V_{SM}$}                 & \multirow{2}{*}{$0.611~(1.488)$}                  & \multirow{2}{*}{$T_{car}~(+0.56)$}                         & \multirow{2}{*}{$\frac{C_{SM}}{T_{car}}~(-0.16)$}          \\\\
    \multirow{2}{*}{$V_{car}$}                & \multirow{2}{*}{$0.268~(1.163)$}                  & \multirow{2}{*}{$\frac{T_{SM}C_{train}}{C_{car}}~(+0.61)$} & \multirow{2}{*}{$\frac{T_{car}}{T_{train}}~(-0.49)$}       \\ \\\bottomrule
  \end{tabular}
\end{table}

\begin{figure}[t]
  \centering
  \includegraphics[width=0.6\linewidth]{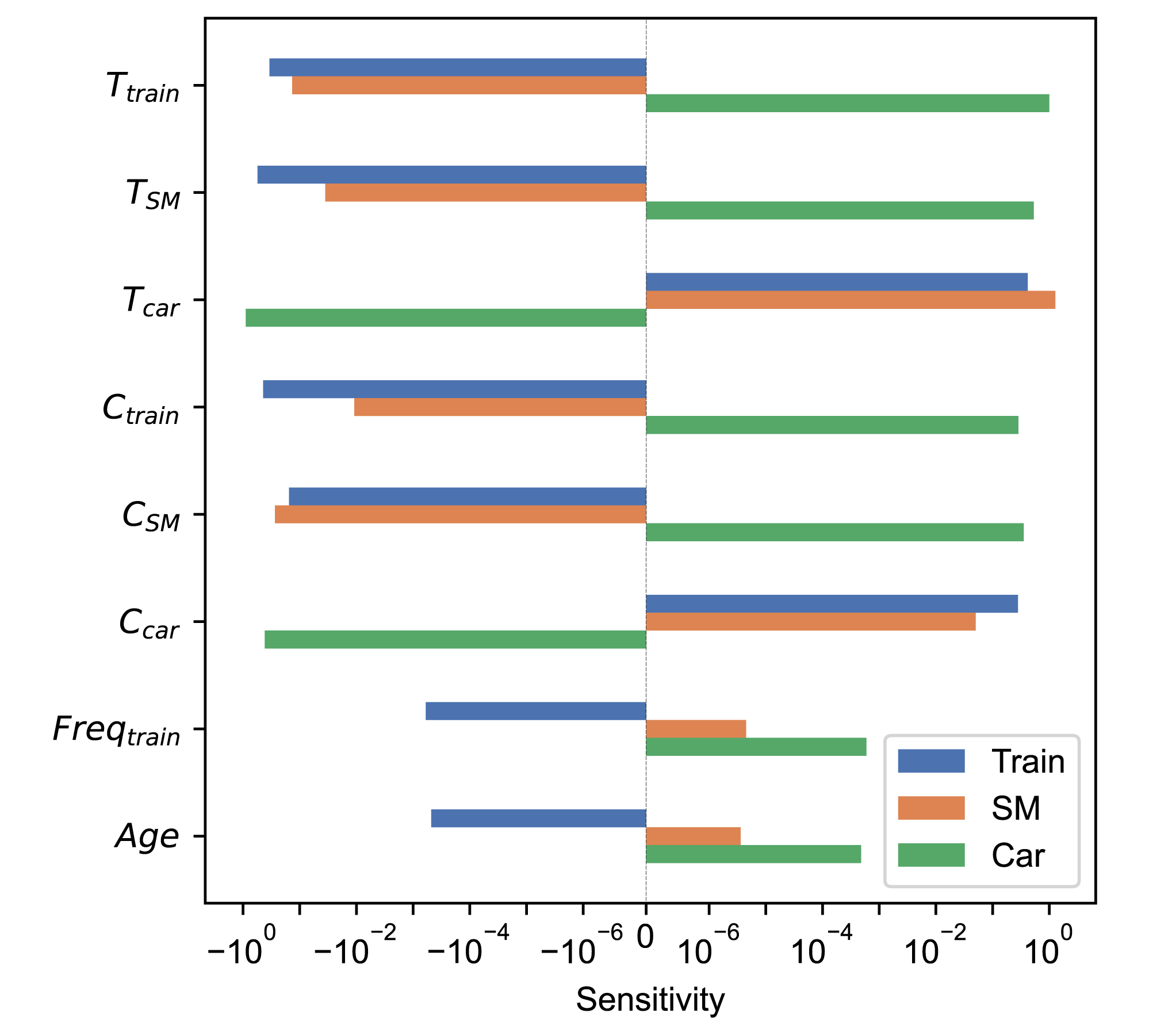}
  \caption{Sensitivity of utility functions to input variables. \rm{Sensitivities are shown using a symmetric log scale with a linear threshold of $1e^{-6}$, which does not affect appearance and interpretation.}}
  \label{fig:sensitivity}
\end{figure}

\subsubsection{Sensitivity analysis}
Although Diff-DCM reveals effective factors that determine the travel mode choice in an interpretable form, for better interpretation and planning interventions, it is helpful to know the effect of a certain variable on the utilities, i.e.~sensitivity.
Such an effect is difficult to distinguish particularly when the utility functions have complex interaction terms, as in the results of this study.
In this case, the sensitivity can be easily obtained via automatic differentiation.
Figure~\ref{fig:sensitivity} shows the calculated sensitivity of utilities of the travel mode to the input variables.
The results show that time and cost have huge impacts on utilities that are orders of magnitude larger than those of the other variables.
Because the sensitivity exhibited similar behaviours for train and SM, it indicates that people see these modes in a similar way.
This result is reasonable considering that, although SM is not a usual train, it is a fast mag-lev underground system similar to a train~\cite{bierlaire_2001,bierlaire_2018}.
Without carefully inspecting the generated terms in utility functions, sensitivity analysis using automatic differentiation can provide information on what kinds of variables have significant effects on choice decisions.

\subsubsection{Prediction performance}
After validating the meaningfulness of the model estimated by Diff-DCM, we confirmed its prediction performance using the test dataset.
In the prediction test, Diff-DCM achieved an accuracy of 67.6\% in predicting travel mode choice, which surpassed the accuracy of the expert-designed model (64.3\%).
Regarding the log-likelihood score $LL$, the performance of Diff-DCM was $LL=-1326.764$, whereas that of the expert-designed model was $LL=-1434.731$, corresponding to an 8\% improvement.
This result verified that Diff-DCM better expresses the actual mode choice tendencies than does the expert-designed model, even without any prior knowledge about the travel mode choice problem.

\section{Discussion}
We introduced Diff-DCM, a fully data-driven but interpretable method for model specification, learning, prediction, and control of human behaviour.
Conventional DCM use expert-designed linear-in-parameter utility functions for model specifications; conversely, Diff-DCM can automatically estimate various forms of utility functions solely from data without any prior knowledge.
With both synthetic and real-world datasets, we validated that Diff-DCM can be applied to various choice dataset.
These estimations could be done in a very efficient way using gradients and completed within 30~s on a laptop without GPUs.
We also demonstrated that gradients can be utilised in various ways such as for sensitivity analysis and optimal intervention planning for behavioural changes.

The requirement of expert knowledge in behavioural model specification has been a long-standing challenge, being a barrier towards fully-automated and interpretable human behaviour modelling.
Utilising differentiable programming and a specific model architecture, Diff-DCM achieved fully data-driven but interpretable behavioural model specification for the first time.
The results of this study ultimately contribute to the automated building of human digital twins~\cite{lin_2024} or live simulations~\cite{swarup_2020}, in which behavioural modelling would be an essential part.
Additionally, the differentiability of Diff-DCM enables it to determine the optimal intervention path towards desired behavioural changes, as demonstrated in this study.
These two features of Diff-DCM pave the way for fully automated behavioural modelling and optimal intervention planning.

There are two primary directions for future research efforts seeking to expand upon this study.
The first is the extension of Diff-DCM.
Although Diff-DCM can be applied to various choice problems, an extension would be required for some problems, such as choice involving nests or hierarchical structures.
To address such problems, conventional DCM offers the nested logit model~\cite{mcfadden_1978}.
The differentiable implementation of such extensions is an essential future work, which will certainly broaden the applicability of Diff-DCM.
The second is integrating Diff-DCM into differentiable agent-based simulations~\cite{andelfinger_2022,chopra_2023,arnau_2023,joel_2023}.
Currently, Diff-DCM alone requires choice outcomes to estimate the model.
The choice observations are easily available in the recent data-rich environment; however, it will be more useful if we can estimate behavioural models from more broader types of observations.
By integrating Diff-DCM into differentiable simulations, simulators that include the interpretable behavioural model will become end-to-end differentiable.
Subsequently, we can compute the gradient of the loss, which is calculated based on simulated behaviours and observations, with respect to parameters in Diff-DCM.
This enables the estimation of the behavioural model from various observations via efficient gradient-based optimisations.
The sampling occurs during the simulation because choices simulated with Diff-DCM are stochastic; however, recently developed techniques, such as the reparameterisation trick~\cite{kingma_2013, jang_2017,gaurav_2022}, enable such simulations with discrete components to be made end-to-end differentiable.
These future works would contribute to the automated building of reliable digital twins of various social phenomena, in which human behaviours are automatically modelled in an interpretable way from various observation data, and information on optimal interventions leading to an improved society can be generated for policy makers.

\bibliographystyle{unsrtnat}

\end{document}